\documentclass[sigconf]{acmart}
\usepackage{multirow}
\usepackage{enumerate}
\AtBeginDocument{%
  \providecommand\BibTeX{{%
    \normalfont B\kern-0.5em{\scshape i\kern-0.25em b}\kern-0.8em\TeX}}}

\copyrightyear{2021} 
\acmYear{2021} 
\setcopyright{acmlicensed}\acmConference[MM '21]{Proceedings of the 29th ACM International Conference on Multimedia}{October 20--24, 2021}{Virtual Event, China}
\acmBooktitle{Proceedings of the 29th ACM International Conference on Multimedia (MM '21), October 20--24, 2021, Virtual Event, China}
\acmPrice{15.00}
\acmDOI{10.1145/3474085.3475520}
\acmISBN{978-1-4503-8651-7/21/10}

\begin{document}


\title{From Image to Imuge: Immunized Image Generation}

\author{Qichao Ying}
\email{shinydotcom@163.com}
\affiliation{%
  \institution{Fudan University}
  \city{Shanghai}
  \country{China}
}

\author{Zhenxing Qian}
\authornote{Corresponding author.}

\email{zxqian@fudan.edu.cn}
\affiliation{%
  \institution{Fudan University}
  \city{Shanghai}
  \country{China}
}

\author{Hang Zhou}
\email{zhouhang2991@gmail.com}
\affiliation{%
  \institution{Simon Fraser University}
  \city{Burnaby}
  \state{British Columbia}
  \country{Canada}
}

\author{Haisheng Xu}
\email{hansonx@nvidia.com}
\affiliation{%
  \institution{NVIDIA}
  \city{Shanghai}
  \country{China}
}

\author{Xinpeng Zhang}
\email{zhangxinpeng@fudan.edu.cn}
\affiliation{%
  \institution{Fudan University}
  \city{Shanghai}
  \country{China}
}

\author{Siyi Li}
\email{louli@nvidia.com}
\affiliation{%
  \institution{NVIDIA}
  \city{Shanghai}
  \country{China}
}

\fancyhead{}


\renewcommand{\shortauthors}{Ying et al.}

\begin{abstract}
We introduce Imuge, an image tamper resilient generative scheme for image self-recovery. The traditional manner of concealing image content within the image are inflexible and fragile to diverse digital attack, i.e. image cropping and JPEG compression. To address this issue, we jointly train a U-Net backboned encoder, a tamper localization network and a decoder for image recovery. Given an original image, the encoder produces a visually indistinguishable immunized image. At the recipient’s side, the verifying network localizes the malicious modifications, and the original content can be approximately recovered by the decoder, despite the presence of the attacks. Several strategies are proposed to boost the training efficiency. We demonstrate that our method can recover the details of the tampered regions with a high quality despite the presence of various kinds of attacks. Comprehensive ablation studies are conducted to validate our network designs.
\end{abstract}

\ccsdesc[500]{Computing methodologies~Image representations}

\keywords{image trust; image restoration; tamper localization}

\begin{teaserfigure}
  \includegraphics[width=0.95\textwidth]{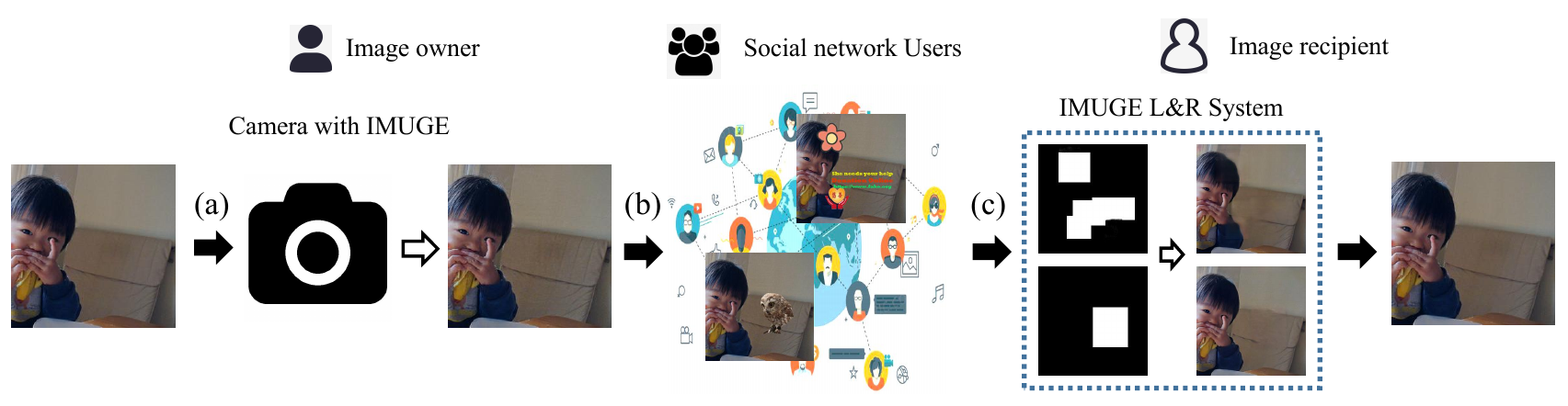}
  \caption{A practical application of the proposed scheme (Imuge). (a) The image owner captures a scene using a camera equipped with Imuge that generates a corresponding immunized image of this scene. (b) The owner uploads the immunized image onto the cloud where the social network users redistribute and modify the image. (c) The recipient with the Imuge L\&R (Localization and Recovery) system can identify the attacks and conduct content recovery.}
  \label{fig:teaser}
\end{teaserfigure}

\maketitle


\section{Introduction}
Digital images are vulnerable to malicious tampers in social networks. The attackers may remove the copyright information, conduct malicious image processing operations or even replace some important contents with fake ones before sending an image on the communication channels. Such a scenario sometimes leads to severe semantic misunderstandings. Therefore, to protect image authenticity as well as ownership, it is essential to design a technique that can automatically recover the image’s original contents against malicious tampers ahead of or after receiving the image.

Previously, many fragile watermarking schemes have been developed for digital images to accurately detect the areas that are modified with counterfeit information~\cite{lu2003fragile,he2006wavelet,zhang2011self,zhang2008fragile,zhang2009fragile,zhang2010reference,zhang2011watermarking,chen2017self,preda2015watermarking,tsai2008authentication}. The original content in the manipulated or missed regions can be recovered after data extraction in the rest of the regions. Despite that these state-of-the-art watermarking schemes for self-recovery can achieve promising content recovery performance, they are not robust to benign attacks. In the real-world applications such as data sharing in the social media, image will suffer from various kinds of distortions, such as lossy compression, image scaling, image enhancement, etc. It is impossible to obtain an uncompressed image to realize self-recovery using these schemes. 

In this paper, we propose a novel generative network called ``Imuge"" to address this issue by generating images immune to tampers. Specifically, we localize the malicious modifications on the immunized image after data sharing, so that the modifications can be reverted, ensuring that the recovered image is approximately identical to the original one.  Figure~\ref{fig:teaser} illustrates a practical application of Imuge. The owner shares the immunized image on the Internet, where the attacker has access to the immunized image and launch malicious tampers. The attacks might destroy the image fidelity or copyright information. During the data transmission, other benign attacks such as compression, scaling and addition of Gaussian noise are introduced. On the recipient's side, the image recipient downloads the attacked image from the Internet, and he/she can conduct tamper localization and image recovery to authenticate the doubted image. There are three main objectives for Imuge. 1) Effectiveness. Imuge localizes the tampered areas in the protected image and successfully conducts content recovery. After content recovery, the recovered image should preserve high visual quality. 2) Robustness. Imuge needs to be tolerant to benign image processing operations or perturbations such as lossy compression, scaling, etc. The robustness against malicious tampers should not deteriorate with the presence of these attacks. 3) Imperceptibility. Imuge works silently and the distortion caused by Imuge is imperceptible.

\begin{figure*}[ht]
\label{img_frame}
	\centering
	\includegraphics[width=0.95\textwidth]{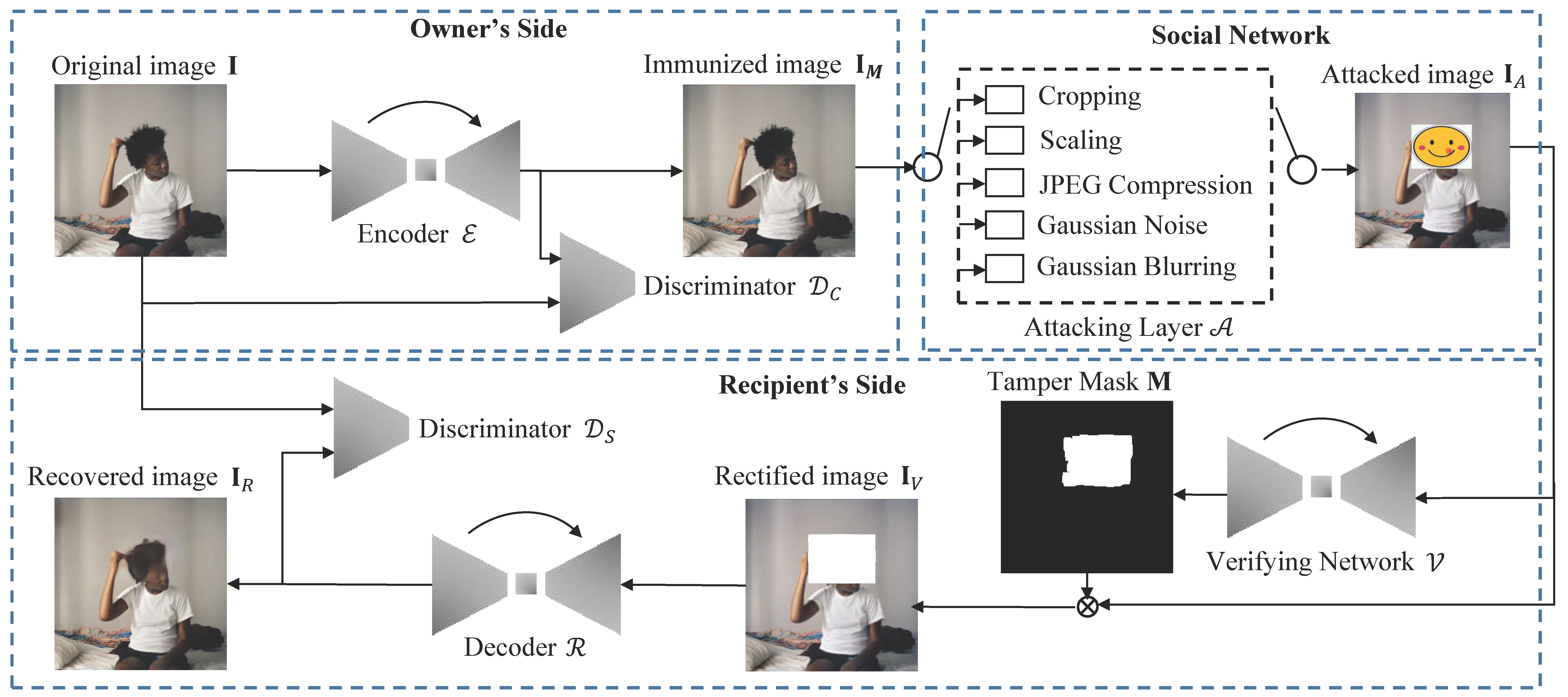}
	\caption{Overview of Imuge network structure. The encoder  generates the immunized image, which is attacked by the attacking layer. The verifying network predicts the tamper mask of the attacked image and generates the rectified image. Finally, the decoder generates the recovered image given the rectified image.}
	\label{fig2}
\end{figure*}

We implement Imuge based on the generative adversarial networks (GAN) ~\cite{goodfellow2014generative}. We jointly train an encoder, a verifying network and a decoder. A parameter-less noise layer is proposed to simulate the attacks. The models are based on self-supervised deep networks, i.e., the procedures are carried out without any manual tagging or assistance. Besides, we also propose several effective training mechanisms, namely, task decoupling, progressive recovery and local feature sharing, to improve the performances of Imuge. In the experiments, we add several man-made real-world attacks on some immunized images and let Imuge localize the tamper and recover the original content. The results show the effectiveness of Imuge in tamper localization and content recovery.

The main contributions of this paper is three-folded, namely:

\begin{enumerate}[1)]
\item We propose the first deep-learning based scheme (Imuge) that realizes robust image tamper localization and content recovery.
\item The immunized images are visually identical to the original images, but modifications made on the immunized images can be removed and the original content can be recovered.
\item High accuracy of tamper localization and good quality \& high fidelity of the recovered image can be achieved.
\end{enumerate}

\section{Related Works}
\label{section:related}
In the past decade, many effective schemes of fragile watermarking for self-recovery have been proposed. In the early methods, He et al. \cite{he2006wavelet} and Lu et al. \cite{lu2003fragile} merely embed the discrete cosine transform (DCT) coefficients or a low color depth version into the least significant bit (LSB) planes. The schemes can only achieve rough content recovery. Zhang et al. \cite{zhang2011self} proposes an improved scheme where the watermark data produced by exclusive-or operation on the original MSB of a pair of pixels are embedded into the LSB planes. The receiver may estimate their values according to the original or recovered neighbor pixels. Later, Zhang et al. further proposes several schemes \cite{zhang2008fragile,zhang2009fragile} where the watermark data is derived from the entire original content and embedded into the host using a reversible data hiding (RDH) technology. Zhang et al. \cite{zhang2010reference} proposes a reference sharing mechanism, in which the watermark to be embedded is a reference derived from the principal content in different regions and shared by these regions for content recovery. Later, Zhang et al. \cite{zhang2011watermarking} proposes a watermarking scheme with flexible recovery quality. It mainly deals with watermark-data waste problems, which is to conduct self-embedding more efficiently. The precision of coefficient recovery is dependent on the amount of available watermark data. 

These schemes have some shared drawbacks. Firstly, they are effective only if the percentage of tampered area is low, e.g., less than 3.2\% in \cite{zhang2008fragile}, and 6.6\% in \cite{zhang2010reference}. However, in real world application, the percent might be much higher, e.g., 20\%. Secondly, whether a block is classified as “tampered” relies on the hash comparison. The schemes are therefore fragile to attacks in that any modification on either the reserved hash code or the rest of the image content will easily let the schemes make wrong judgements. Though there are some further improved schemes \cite{preda2015watermarking,tsai2008authentication} that generate robust hash codes for the blocks, they are only tolerant to a typical kind of attack such as JPEG compression but still sensitive to the other common attacks. In summary, it still remains a big issue to develop a novel scheme for robust image self-recovery.

\section{Proposed Scheme}
\label{section:method}
\subsection{The Overall Pipeline}

Motivated by the shortcomings of existing methods, we propose a novel image tamper resilient generative scheme (Imuge) to construct the immunized images that are immune to the tamper attacks. Figure~\ref{fig2} shows an overview network architecture of Imuge. Our goal is to convert normal images into immunized images and to conduct successful tamper localization and content recovery on them at the recipient’s side. Imuge is designed to be robust against common attacks such as lossy compression, image interpolation or cropping. Imuge begins with slightly modifying the original image $\mathbf{I}$ using the encoder $\mathcal{E}$ to produce the immunized image $\mathbf{I}_M$. Then, the attacking layer $\mathcal{A}$ applies different kinds of simulated attacks on $\mathbf{I}_M$ and produces the attacked image $\mathbf{I}_A$. The verifying network $\mathcal{V}$ receives the attacked image and localizes the tampered areas by predicting a tamper mask $\mathcal{M}$, which pixel-wisely classifies whether each pixel in $\mathbf{I}_A$ is tampered. Then we generate the rectified image $\mathbf{I}_V$ by removing the tampered contents within $\mathbf{I}_A$ according to the predicted mask. Finally, the decoder $\mathcal{R}$ receives $\mathbf{I}_V$ and produces the recovered image $\mathbf{I}_R$, where the missing contents are recovered. We further introduce two discriminators, namely, the discriminator $\mathcal{D}_C$ and the discriminator $\mathcal{D}_S$ to improve the image quality of the immunized image $\mathbf{I}_M$ and the recovered image $\mathbf{I}_R$. The discriminators distinguish the generated images respectively from the original image $\mathbf{I}$.

\subsection{Network Architecture}
In many image processing tasks with deep neural networks, the U-Net \cite{ronneberger2015u} has been widely used as the baseline architecture. The skip connections can preserve low level information and provide a massive receptive field. The backbone U-Net architecture in Imuge is constructed as follows: The encoding part of the U-Net is consist of four encoding segments, and each segment increases the number of channels by a factor of two. The segment include the convolutional layers, the normalization layers and the activation units. The spatial size of the features is reduced using a pooling layer with stride-2. The size of the output features of the encoding part is therefore $1/16 \times 1/16$ of the original image. The decoding part is consist of four decoding segments, and each segment expands the spatial size of the input and reduces the number of channels by a factor of two. The segments include the transposed convolutional layers with stride-2, the normalization layers and the activation units. The features of the same level from the encoding and the decoding part are concatenated before we send them into the next decoding segment. After the last segment of the decoder part, the final output is generated by applying a $1 \times 1$ convolutional layer followed by a tanh activation. Between the encoder part and the decoder part, there is an additional convolutional segment which uses dilated convolutional layers \cite{yu2015multi}. 

The encoder $\mathcal{E}$ is implemented by the backbone architecture. We accept the residual learning strategy on generating the immunized image $\mathbf{I}_M$. The network outputs a residual image $\mathbf{R}$ and add it onto the original image $\mathbf{I}$ to generate the immunized image $\mathbf{I}_M$.

In the attacking layer $\mathcal{A}$, we simulate several types of common attacks. In Imuge, typical image attacks are categorized into the malicious tampers and the benign attacks. The layer implements five types of attacks and two types of malicious tampers. The attacks include: 1) Adding white Gaussian noise. We introduce additive white gaussian noise (AWGN) on the immunized image $\mathbf{I}_{M}$. The mean and the standard deviation of AWGN respectively are $\mu=0$, $\sigma=0.1$. 2) Gaussian blurring. We filter the immunized image $\mathbf{I}_{M}$ with a convolutional layer with kernel size $k \in [3,5]$. 3) Scaling. We randomly scale down or up the immunized image $\mathbf{I}_{M}$, and scale it back to the original size. The ratio of the scaling belongs to $[0.5,2]$. 4) JPEG compression. We apply JPEG compression on the immunized image $\mathbf{I}_{M}$ using a differentiable implementation proposed in \cite{zhu2018hidden}

The malicious tamper selects a number of random areas in $\mathbf{I}_M$ and conduct the following attacks:1) Content replacement. We tamper some random areas of $\mathbf{I}_M$ using the contents of another irrelevant image. Or the areas are directly filled with a single color. 2) Clone Stamp. We tamper some random areas of $\mathbf{I}_M$ by referring to their ambient areas. For simplicity, here we use a spatially shifted version of $\mathbf{I}_M$ to conduct the content replacement.

In both ways, the attacked image $\mathbf{I}_{A}$ is generated by (\ref{eqn_irr}), where $\mathbf{I}_{irr}$  refers to the irrelevant tamper. The arithmetic $\cdot$ is pixel-wise multiplication. $\mathbf{M}_R$, the ground truth tamper mask, is a sparse binary matrix in which “1” indicates conducting pixel replacement in the corresponding location.
\begin{equation}
\label{eqn_irr}
    \mathbf{I}_{A}=\mathbf{I}_{irr} \cdot \mathbf{M}_R+\mathbf{I}_{M} \cdot (1-\mathbf{M}_{R}),\\
\end{equation}

\begin{figure}[ht]
	\centering
	\includegraphics[width=0.45\textwidth]{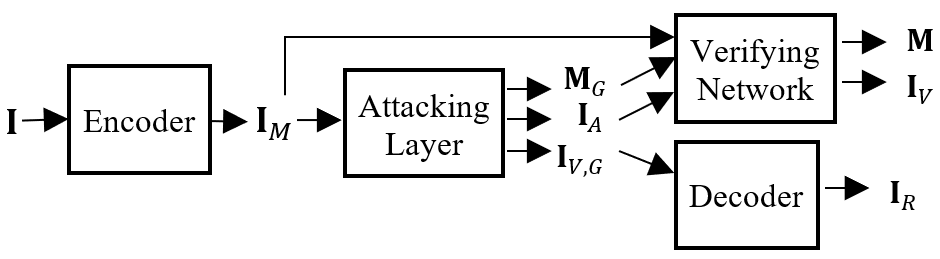}
	\caption{Data flow during task decoupling, where the verifying network is trained in parallel with the decoder. For simplicity, the discriminators are not shown.}
	\label{fig4}
\end{figure}
\begin{figure}[ht]
	\centering
	\includegraphics[width=0.45\textwidth]{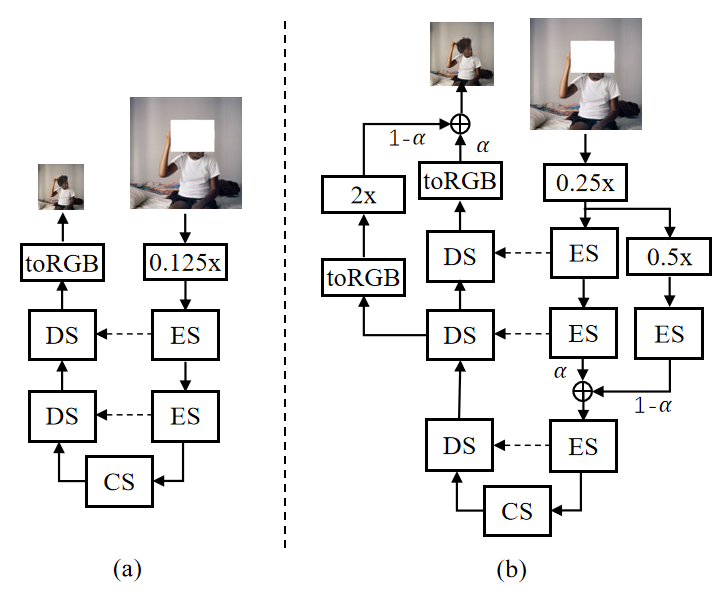}
	\caption{The progressive growing of the decoder $\mathcal{R}$ with progressive recovery. (a) The first stage produces the lowest-resolution recovered images. (b) The second stage generates higher-resolution images. ES/ CS/ DS respectively: the encoding/ convolutional/ decoding segment.}
	\label{image_progress}
\end{figure}
\begin{figure}[ht]
	\centering
	\includegraphics[width=0.45\textwidth]{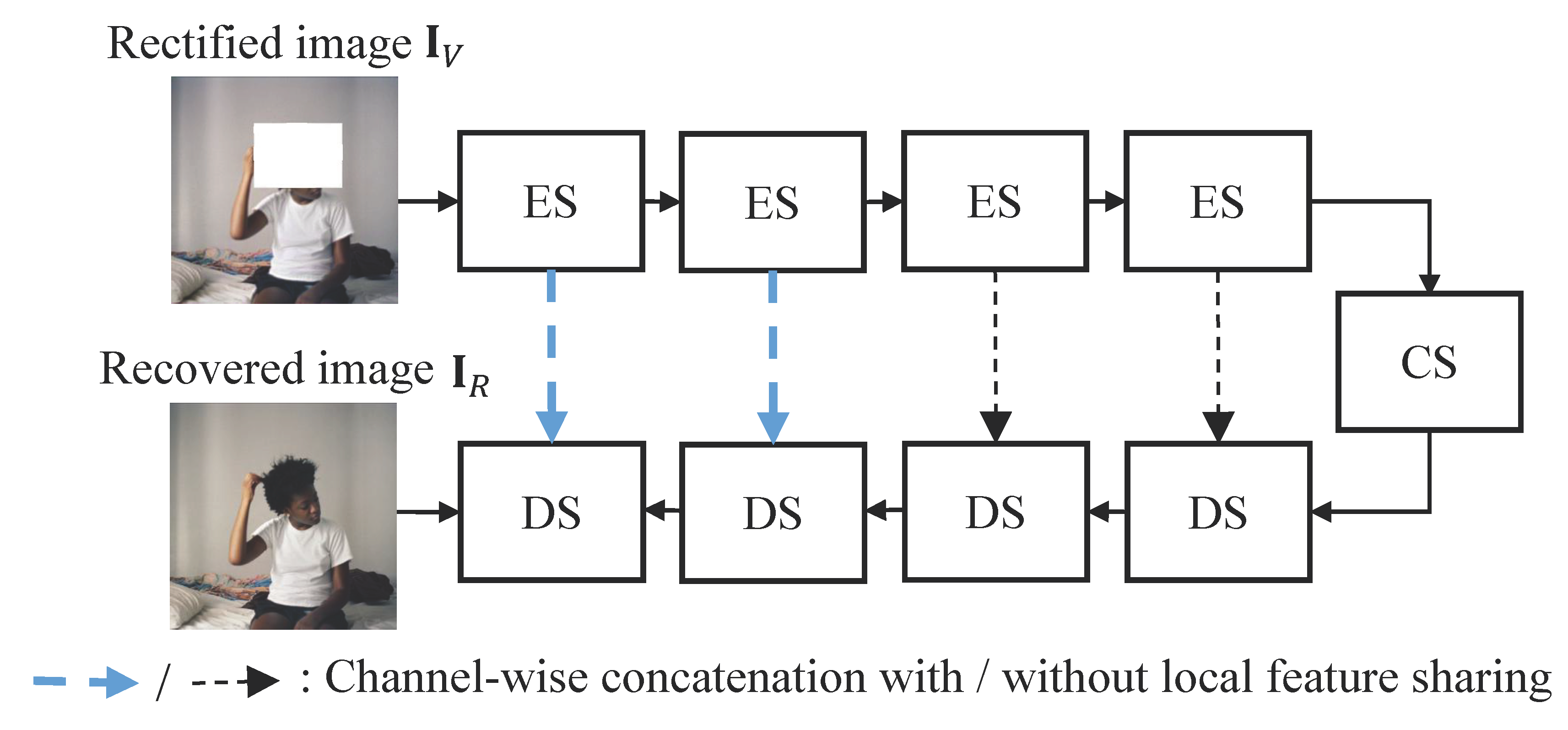}
	\caption{Data flow of the decoder $\mathcal{R}$ by local feature sharing. The mechanism only exists in the first two levels.}
	\label{fig5}
\end{figure}

During training, we always arbitrarily and evenly add one kind of benign attacks and one kind of malicious tamper on the immunized image. Therefore, the networks are evenly trained under various kinds of attacks. The attacks are always added after the tampers. We also arbitrarily skip the attacking layer $\mathcal{A}$ so that the attacked image $\mathbf{I}_A$ is identical to the immunized image $\mathbf{I}_M$. When $\mathcal{A}$ is skipped, we force the verifying network $\mathcal{V}$ to predict a zero matrix to represent that there is no tamper.

The verifying network $\mathcal{V}$ and the decoder $\mathcal{R}$ also accept the backbone architecture as the implementation, where $\mathcal{V}$ produces a single-channeled output . Finally, we use the Patch-GAN \cite{isola2017image} as the discriminators ($\mathcal{D}_C$ and $\mathcal{D}_S$).

\subsection{Mechanisms for Better Performance}
\label{section_technique}
Successful content recovery from the tampered image is a difficult task in that: not only is the distribution of the location of tampered area within an immunized image close to random, but the size of the tampered areas also vary in the real-world application. In this section, we propose several useful training mechanisms that improve the performance, namely, task decoupling, progressive recovery and local feature sharing.

\subsubsection{Task Decoupling.}
In the real-world applications, content recovery can only be conducted after tamper localization. If tamper localization fails, the wrongly predicted rectified image $\mathbf{I}_V$ will mislead the decoder, which usually will lead to failed content recovery. But the failure of precise tamper mask prediction always happens when the networks of the Imuge are under-fitting. If the two tasks are coupled at the beginning of training, the decoder $\mathcal{R}$ will have to conduct content recovery over a wrongly predicted tamper mask. As a result, none of the networks can be properly trained. 

To address this issue, we introduce task decoupling during the training stage. Figure~\ref{fig4} shows the data flow during the task decoupling stage. When the tamper classification loss of $\mathcal{V}$ (detailed in Section \ref{subsection:lossfunc}) does not gradually converge, we directly provide the decoder $\mathcal{R}$ with the ground-truth masked image $\mathbf{I}_{V,G}$ of the attacked image $\mathbf{I}_A$. In $\mathbf{I}_{V,G}$, the contents within the tampered areas are removed from the attacked image, i.e., $\mathbf{I}_{V,G}= \mathbf{I}_A \cdot \mathbf{M}_G$. Besides, we train the verifying network using $\mathbf{I}_A\}$ and $\mathbf{M}_G$. During task decoupling, the verifying network $\mathcal{V}$ is trained in parallel with the decoder $\mathcal{R}$, and the outputs of $\mathcal{V}$ is not used to produce the recovered image.

Task decoupling allows the verifying network $\mathcal{V}$ to temporarily focus on the learning of providing precise tamper mask prediction, while the rest of the networks concentrate on providing robust image immunization. We lift the task decoupling when the tamper classification loss is gradually converged, or after an empirical number of epoches. After task decoupling, the data flow of Imuge is identical to that in the infering stage, where the recovered image $\mathbf{I}_{R}$ is generated according to the predicted tamper mask provided by the verifying network, i.e., $\mathbf{I}_{V}=\mathbf{I}_A \cdot \mathbf{M}$.

\subsubsection{Progressive Recovery.}
Inspired by PG-GAN \cite{karras2017progressive}, we conduct the content recovery by reconstructing the recovered image from lower resolution to higher resolution. The progressive image reconstruction divides content recovery into several stages. We propose a progressive reconstruction strategy for the U-Net architectures.

Figure~\ref{image_progress} shows the first two stages of the proposed progressive recovery mechanism. In the first stage, we first down-sample the attacked image to $1/8 \times 1/8$ the size of the original image. We feed the down-sampled image into the fourth-leveled encoding segment, and the low-resolution recovered image is generated by the fourth-leveled decoding segment. The three upper-leveled encoding segments and decoding segments are not used for image generation. In the second stage, we involve the third-leveled encoding segment and the third-leveled decoding segment to generate the recovery image. The size of the recovered image at this stage is $1/4 \times 1/4$ of the original image. The previous existing layers remain trainable. Both the newly included segments from the encoding part and the decoding part are faded in smoothly by a parameter $\alpha$ to avoid sudden attack to the previous layers. We continue the progressive recovery stages until the decoder $\mathcal{R}$ produces the recovered image $\mathbf{I}_M$ whose size is identical to the original image $\mathbf{I}$. Therefore, there are four stages for Imuge to finish the progressive recovery. The discriminator $\mathbf{D}_S$ distinguish the recovered image and the original images at each stage. The progressive growing of the discriminator can be easily inferred from PG-GAN \cite{karras2017progressive}. Notice that the verifier network $\mathcal{V}$ is idle when the progressive recovery mechanism does not reach its last stage, i.e, $\mathcal{V}$ only makes prediction on the full-sized recovered image. Therefore, before the decoder produces the full-sized recovered image, we can regard that Imuge is also task decoupled except that $\mathcal{V}$ is not trained.

\subsubsection{Local Feature Sharing. }
The features generated by the decoding segments in the U-Net based decoder $\mathcal{R}$ have come through more convolutions compared to those generated by the encoding segments. We observe that the features generated by the decoding segments carry richer information of the original content within the tampered area. Thus, we propose the local feature sharing mechanism to share the information within the vacated area between the first two levels of the segments. We denote respectively the output of the third-leveled decoding segment and that of the second-leveled encoding segment as $F_{D,3}$ and $F_{E,2}$. Then, local feature sharing generates a new feature $F_{G,2}$ to combine $F_{D,3}$ with $F_{E,2}$ by:
\begin{equation}
\label{eqn:feat_share}
    F_{G,2} = F_{D,3}\cdot \mathbf{M}_{G,3}+F_{E,3} \cdot(1-\mathbf{M}_{G,3}),\\
\end{equation}
where $\mathbf{M}_{G,3}$ is the down-sampled version of $\mathbf{M}$. The input of the second-leveled decoding segment, denoted as $F_{D,2}$ is therefore $F_{D,2} = \textrm{Concat}[F_{G,2},F_{D,3}]$, where ``Concat'' represents channel-wise concatenation. The local feature sharing in the first level can be easily inferred from (\ref{eqn:feat_share}). In upper levels, we do not apply the feature sharing mechanism.

\subsection{Objective Loss Function}
\label{subsection:lossfunc}

The objective loss function of Imuge consists of three parts: the tamper classification loss, the image reconstruction loss, and the discriminative loss. Formally:
\begin{equation}
\label{eqn_sum}
\mathcal{L}_{\mathcal{G}}=\mathcal{L}_{rec}+\alpha\mathcal{L}_{cls}+\beta\mathcal{L}_{dis},
\end{equation}
where $\alpha$, $\beta$ are weight factors.

$\mathcal{L}_{cls}$ is the tamper classification loss that urges the verifying network $\mathcal{V}$ to make correct prediction to the ground-truth tamper mask $\mathbf{M}_G$. We use the binary cross entropy (BCE) loss between the estimated tamper mask $\mathbf{M}$ and the ground-truth $\mathbf{M}_G$.
\begin{equation}
\label{eqn_class}
\mathcal{L}_{cls}=-[\mathbf{M}\log{\mathbf{M}_G}+(1-{\mathbf{M}}\log{(1-\mathbf{M}_G})].
\end{equation}

$\mathcal{L}_{rec}$ is the reconstruction loss that encourages the immunized image $\mathbf{I}_M$ and the recovered image $\mathbf{I}_R$ to respectively resemble the original image $\mathbf{I}$. We adopt the $\ell_2$ distance as the criterion of the reconstruction. The reconstruction loss for the recovered image $\mathcal{L}_R$ is defined as:
\begin{equation}
\label{eqn:recon_recover}
    \mathcal{L}_R=\|\mathbf{I}-\mathbf{I}_R\|_2^2+\|\mathbf{I} \cdot \mathbf{M}_G-\mathbf{I}_R \cdot \mathbf{M}_G\|_2^2/\|\mathbf{M}_G\|_1.\\
\end{equation}

The second part of (\ref{eqn:recon_recover}) is to put additional emphasis on the reconstruction of content within the tampered areas. The reconstruction loss for the recovered image $\mathcal{L}_C$ is the $\ell_2$ distance between the original image and the immunized image. $\mathcal{L}_C=\|\mathbf{I}-\mathbf{I}_M\|_2^2$. Therefore,
\begin{equation}
\mathcal{L}_{rec}=\mathcal{L}_R+\gamma\mathcal{L}_C,
\end{equation}
where $\gamma$ is the weight factor that balances the fidelity of the immunized image and the degree of image recovery. According to our measurement, we find that the $\ell_2$ distance is better than the $\ell_1$ distance and the Charbonnier distance.

$\mathcal{L}_{dis}$ is the adversarial loss that aim to fool the adversarial networks to make wrong predictions on whether an image is an original image $\mathbf{P}$ or a generated image $\hat{\mathbf{P}}$.  We accept the least squared adversarial loss proposed in \cite{mao2017least}, which is defined as: 
\begin{equation}
\mathcal{L}_{dis}(\hat{\mathbf{P}})=\lVert 1-\mathcal{D}_{\theta}(\hat{\mathbf{P}}) \rVert_2^2.
\end{equation}

The adversarial loss for $\mathbf{I}_{M}$ and $\mathbf{I}_{R}$ are respectively $L_{DC}=\mathcal{L}_{dis}(\mathbf{I}_{M})$ and $L_{DR}=\mathcal{L}_{dis}(\mathbf{I}_{R})$. Therefore,
\begin{equation}
\label{eqn:1t}
\mathcal{L}_{dis}=\mathcal{L}_{DR}+\theta\mathcal{L}_{DC},
\end{equation}
where $\theta$ is the weight factor that has the same goal with $\gamma$.


The losses of the two adversarial networks aim respectively to distinguish real and fake images by minimizing:
\begin{equation}
\label{eqn:1v}
\mathcal{L}_{\mathcal{D}}(\mathcal{P},\hat{\mathcal{P}})=\frac{1}{2}\lVert \mathcal{D}_{\theta}(\hat{\mathcal{P}}) \rVert_2^2+\frac{1}{2}\lVert 1-\mathcal{D}_{\theta}(\mathcal{P}) \rVert_2^2.
\end{equation}

The discriminator  $\mathcal{D}_{C}$ minimizes $\mathcal{L}_{\mathcal{D}}(\mathbf{I},\mathbf{I}_{M})$, and the discriminator  $\mathcal{D}_{S}$ minimizes $\mathcal{L}_{\mathcal{D}}(\mathbf{I},\mathbf{I}_{R})$.

\subsection{Implementation Details}

The design of the training stage is closely related to the proposed training mechanisms. As stated in Section \ref{section_technique}, we begin with training the networks according to the progressive recovery strategy. Afterwards, task decoupling is applied until the tamper classification loss $\mathcal{L}_{cls}$ gradually converges.

During task decoupling stage, we iteratively train the verifying network $\mathcal{V}$ and the rest of the networks. After the attacking layer $\mathcal{A}$ generates the attacked image $\mathbf{I}_{A}$, the verifying network $\mathcal{V}$ generates the predicted tamper mask $\mathbf{M}$ and only update itself by minimizing $\mathcal{L}_{cls}$ in (\ref{eqn_class}). On the other hand, we provide the decoder $\mathcal{R}$ with the ground-truth rectified image $\mathbf{I}_{V,G}$ and $\mathcal{R}$ generates the recovered image $\mathbf{I}_{R}$. We then update the networks by minimizing the overall loss $\mathcal{L}_{\mathcal{G}}$ in (\ref{eqn_sum}), where $alpha$ equals to zero and the verifying network $\mathcal{V}$ is not trained. When the task decoupling is lifted, the networks are trained in one go. The decoder $\mathcal{R}$ generates the recovered image $\mathbf{I}_{R}$ according to the predicted tamper mask $\mathbf{M}$ provided by the verifying network $\mathcal{V}$, and all the networks are updated by minimizing the overall loss $\mathcal{L}_{\mathcal{G}}$.

In the inferring stage, to avoid edge-blurring effect, we use Otsu's method \cite{otsu1979threshold} to convert $\mathbf{M}$ into a binary matrix. We also slightly enlarge the predicted tampered area using image eroding and dilating operation with kernel size $k=4$ on $\mathbf{M}$. Such a strategy helps to remove incorrect prediction pulses in the real-world application.

\begin{figure}[ht]
	\centering
	\includegraphics[width=0.45\textwidth]{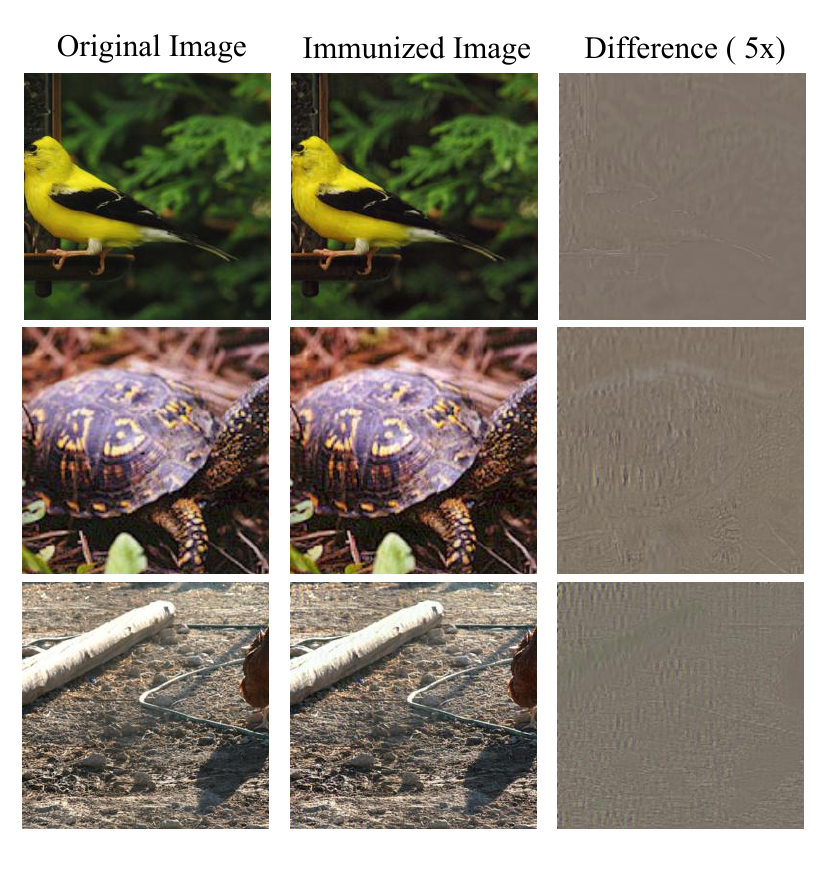}
	\caption{Illustration of three original images and their corresponding immunized images.}
	\label{img_marked}
\end{figure}

\section{Experiments}
\label{section:experiment}

In this section, we provide a comprehensive performance analysis of Imuge in real-world application, where the immunized images are randomly attacked by the volunteers. We also conduct the ablation tests to highlight the pipeline design. It is well noticing that, the fragile self-recovery schemes \cite{lu2003fragile,he2006wavelet,zhang2011self,zhang2008fragile,zhang2009fragile,zhang2010reference,zhang2011watermarking,chen2017self,preda2015watermarking,tsai2008authentication} cannot resist benign attacks such as JPEG compression or scaling. For example, the immunized image will be regarded as fully tampered when it is lossily compressed and therefore the self-recovery procedures cannot be carried out. In contrast, Imuge is designed to conduct image restoration against benign attacks. Therefore, we do not compare Imuge with the previous fragile image self-recovery schemes.

\subsection{Experimental Settings}

During training, the hyper-parameters are set as $\alpha=0.01, \beta=0.005,\gamma=0.5, \theta=0.5$. We train Imuge on two GeForce 1080Ti GPU. For gradient descent, we use Adam optimizer \cite{kingma2014adam} with the default hyper-parameters. The learning rate is $2\times10^{-4}$. All models are trained with batch size 8. The architecture of Imuge is trained for 200 epochs where each phase expect the last of the progressive recovery takes 20 epoches. The verifying network is decoupled from the rest of Imuge at the beginning and the task decoupling lifts after 100 epochs. The training finishes in roughly two weeks. To enable practical usage, we train Imuge on the ILSVRC2017 training set \cite{ILSVRC}, containing 456,567 natural images. We evaluate our model on a mixture of images from different datasets including ILSVRC2017 test set and COCO training set \cite{lin2014microsoft}. The mixture contains 51,000 images that do not occur in the training set. We resize the original images to the size of $256 \times 256$. Imuge is trained/validated with automatically generated attacks and tested with hundreds of human-participated attacks, thanks to the anonymous volunteers. The generated immunized images are saved in BMP format and the attackers save the attacked versions randomly in common formats such as BMP, PNG, TIFF or JPEG.

\subsection{Performance in Real-World Scenario}
\subsubsection{Quality of the Immunized Images.}
In Figure~\ref{img_marked}, we randomly generate immunized images using three randomly-selected test images from the above-mentioned mixture of COCO and ILSVRC2017 datasets. The third column in each group shows the difference between the original image and the immunized image, which is also the residual image $\mathbf{R}$ generated by the encoder $\mathcal{E}$. Since the residual image will contain negative values, they are normalized in the figures for clear illustration. The residual image is generally composed of two patterns judging from the figures. Firstly, it comprises a coarse representation of the original image where content information is widely spread into the ambient areas. This part of the residual image is crucial for content recovery. Secondly, due to the fact that plain areas lack the first kind of residual image, the classification error from the verifying network $\mathcal{V}$ forces the encoder $\mathcal{E}$ to produce the second kind of pattern, which helps to localize the tamper attacks. Th latter pattern explains the necessity of lifting the task decoupling at the end of training. The addition of the residual image $\mathbf{R}$ is imperceptible, but can be recognized by the verifying network $\mathcal{V}$ and utilized by the decoder $\mathcal{R}$ to conduct content recovery. The average PSNR between the immunized images and the original images in Figure~\ref{img_marked} is 34.60dB, and the average SSIM \cite{wang2004image} is 0.973. 

\begin{figure*}[ht]
	\centering
	\includegraphics[width=0.95\textwidth]{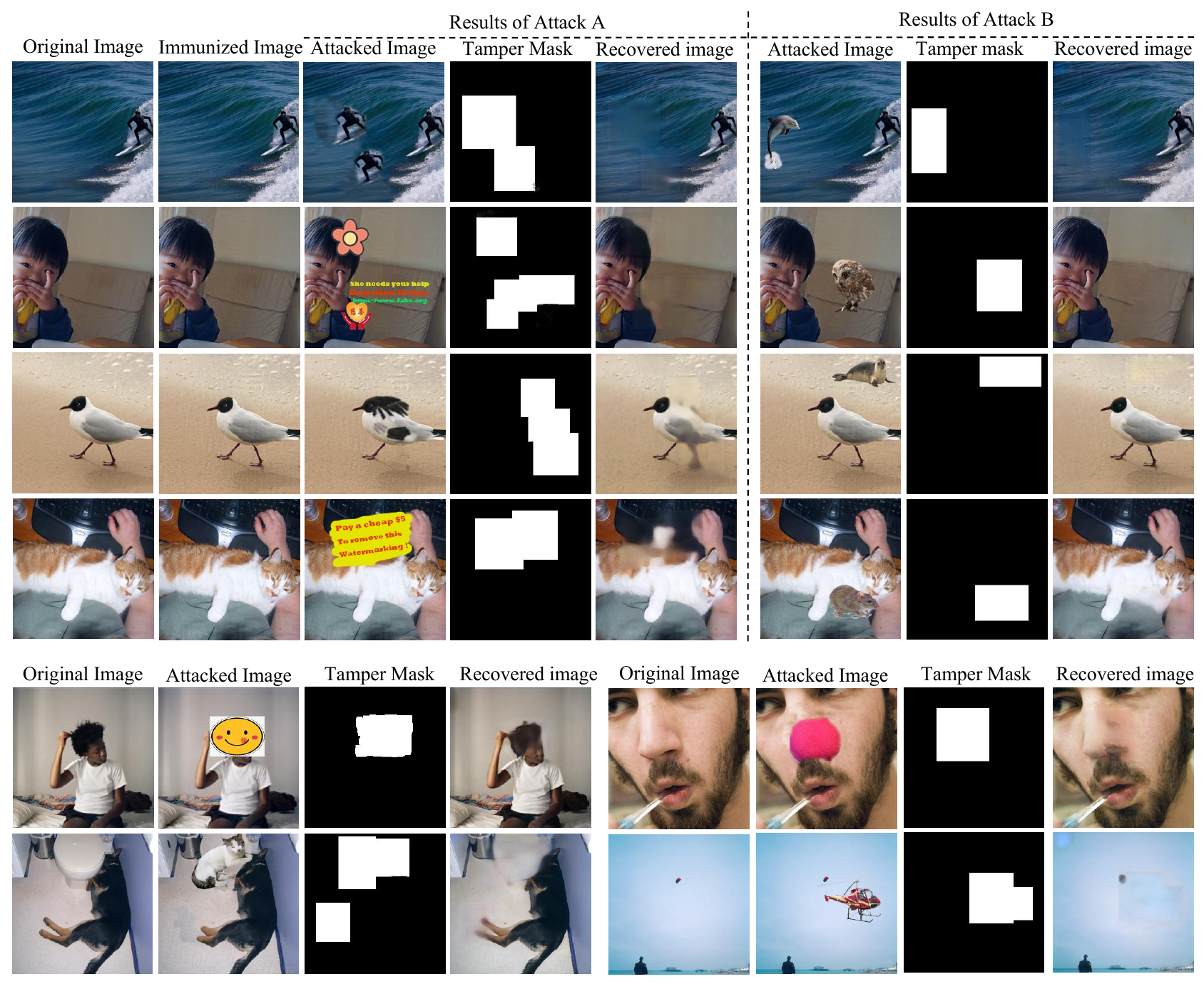}
	\caption{Examples of tamper localization and content recovery. The tampered images are saved randomly in JPEG format.}
	\label{fig7}
\end{figure*}
\begin{table*}[!t]
	\caption{Average performance of tamper localization and image recovery over the 134 immunized images.}
	\label{table_comparison}
	\begin{center}
	\begin{tabular}{c|ccc|ccc|ccc|cccc}
		\hline\toprule[0.4pt]
		\multirow{2}{*}{Index} & \multicolumn{3}{c|}{JPEG} & \multicolumn{3}{c|}{Scaling} & \multicolumn{3}{c|}{Cropping} & \multirow{2}{*}{Blurring} & \multirow{2}{*}{AWGN}& \multirow{2}{*}{No attack} \\
		& QF=90 & QF=70 & QF=50 & 150\% & 70\% & 50\% & 90\% & 70\% & 50\% \\
        \midrule
        BCE & 0.022 & 0.096 & 0.106 & 0.148 & 0.074 & 0.121 & 0.027 & 0.082 & 0.155 & 0.052 & 0.103 & 0.015\\
        L-PSNR & 26.11 & 23.85 & 18.71 & 23.87 & 25.10 & 24.52 & 26.85 & 24.73 & 24.10  & 19.93 & 27.06 & 27.11\\
        PSNR & 29.33 & 27.43 & 25.98 & 25.23 & 26.61 & 25.78 & 28.66 & 25.94 & 24.55 & 27.35 & 28.52 & 29.60\\
        SSIM & 0.862 & 0.854  & 0.803 & 0.838 & 0.870 & 0.829 & 0.863 & 0.821 & 0.797 & 0.817 & 0.842 & 0.865 \\
		\hline\toprule[0.4pt]
	\end{tabular}
	\end{center}
\end{table*}

\subsubsection{Human-Participated Real-World Attacks.}
To further prove the effectiveness of Imuge in practical application, we apply different kinds of combined attacks on 134 immunized images using typical image processing tools like Adobe Photoshop by the volunteers. The rate of the summation of the tampered area versus the whole image is roughly $RST \in [0.1,0.5)$, and the rate of the area of the largest tamper versus the whole image $RLT \in [0.1,0.25)$. The manual attacks mainly include: (1) adding, modifying or deleting some important contents in the immunized images which might leads to semantic misunderstanding, (2) adjusting the size of the image (3) saving the modified images in lossy formats. The JPEG quality factor is set among $[50,90]$. Figure~\ref{fig7} shows eight groups of content recovery test in real application, where the first four groups are tested twice with different tamper attacks. The average tamper localization and image recovery performances over the 134 immunized images are reported in Table~\ref{table_comparison}. ``L-PSNR'' denotes the local PSNR within the tampered areas. 

\subsubsection{Accuracy of Tamper Localization.}
The two columns "Tamper mask" in Figure~\ref{fig7} illustrate the predicted tamper masks of the examples. For tamper localization, if a true negative instance appears, Imuge will fail to conduct content recovery in that area. If a false positive instance happens, Imuge will delete the contents within that area and reconstruct them using content recovery. In both case, the recovered image will downgrade in its quality. We observe that although the images are tampered by a variety of random attacks, we succeed in localizing the tampered areas. Judging from the figures, we believe that the verifying network can conduct accurate classification in the actual application. The verifying network $\mathcal{V}$ is believed to localize the tampered areas by finding the regions that does not contain the residual image introduced by the encoder $\mathcal{E}$. Table~\ref{table_comparison} also proves that the immunization can resist benign attacks with the provided BCE losses generally below 0.1.

\subsubsection{Quality of Image Content Recovery.}
From Figure~\ref{fig7}, we can observe that Imuge can conduct successful content recovery given various kinds of tampers. For example, the added objects can be removed and the missing objects can be restored. In addition, we witness that the immunized image is sometimes a little bit different in luminance compared to the original image. Such modification can be effectively reverted by the decoder $\mathcal{R}$. In Table~\ref{table_comparison}, the local PSNR is generally close to PSNR over the whole image when the attacks are not strong. The average SSIM of every test group is above 0.8 and the average local PSNR is around 25. In Table~\ref{Table2}, we study the effectiveness of Imuge given different combinations of $RST$ and $RLT$. The results shows that different $RST$ and $RLT$ has little effect on the content recovery effect. Larger sized tampers may cause the recovery effect to decrease. Therefore, we believe that Imuge is effective in human-participated applications.


\begin{table}[!t]
	\renewcommand{\arraystretch}{1.3}
	\caption{Average performance under different degree of tamper (without benign attacks). RLT: rate of the area of the largest tamper versus the whole image.}
	\label{Table2}
	\centering
	\begin{tabular}{c|cccc}
		\hline\toprule[0.4pt]
		 & BCE & L-PSNR & PSNR & SSIM \\
        \midrule
        $RST\approx10\%, RLT\approx6\%$ & {0.008} & {28.93} & {29.26} & {0.875} \\
        \midrule
        $RST\approx25\%, RLT\approx6\%$ & 0.007 & 25.32 & 24.83 & 0.848 \\
        \midrule
        $RST\approx10\%, RLT\approx16\%$ & 0.021 & 27.25 & 26.94 & 0.863 \\
        \midrule
        $RST\approx30\%, RLT\approx16\%$ & 0.057 & 16.92 & 26.32 & 0.810 \\
		\hline\toprule[0.4pt]
	\end{tabular}
\end{table}
\begin{table}[!t]
	\renewcommand{\arraystretch}{1.3}
	\caption{Ablation study over the whole scheme with JPEG compression ($QF=70$). TD: Task decoupling. PR: progressive recovery. LFS: local feature sharing.}
	\label{Table3}
	\centering
	\begin{tabular}{c|c}
		\hline\toprule[0.4pt]
		Tests & L-PSNR \\
		\midrule
        No $\mathcal{V}$, no discriminators, no TD,PR,LFS& {13.16}\\
        \midrule
        No discriminators, no TD,PR,LFS & {20.99}\\
        \midrule
        No TD,PR,LFS & {23.30} \\
        \midrule
        No PR,LFS & {25.19} \\
		\midrule
		No LFS & {27.03} \\
		\midrule
		Full implementation & {\textbf{27.43}} \\
		\hline\toprule[0.4pt]
	\end{tabular}
\end{table}

\subsection{Imuge Content Recovery vs Inpainting}
Image inpainting \cite{yu2018generative,wang2019musical,iizuka2017globally,nazeri2019edgeconnect} is a famous image restoration technology that restores the missing content by refering to the outer region. The results of inpainting can be natural yet faulty compared to the ground truth in that providing visually pleasing results is the priority other than the reversibility. In contrast, Imuge content recovery highlights fidelity with the help of image immunization. To better benchmark whether the recovered image of Imuge looks like the original image, we conduct experimental comparisons to prove the effectiveness of Imuge.

First, in our human visual quality analysis, we request 20 participants to score over the recovered images generated by Imuge. We also invite the volunteers to score over the recovered images produced by MUSICAL \cite{wang2019musical}, a state-of-the-art inpainting scheme using multi-scaled attention. We skip the tamper localization in Imuge and directly let the two algorithms conduct image recovery towards the same rectified images. The user study involve two terms: subjective visual quality (SVQ) and subjective image fidelity (SIF). In each term, we ask the volunteers to score from zero to five, where a higher score represents better visual quality or image fidelity. The average scores of Imuge content recovery on SVQ and SIF are respectively 4.2 and 4.6, while those of \cite{wang2019musical} 4.6 and 4.0. The results suggest that the recovered contents are more reliable and trustworthy than those generated by \cite{wang2019musical}. 

Second, we manually remove some complete items from the testing images. In the recovered images of Imuge, the average L-PSNR of Imuge is 25.13dB, and that of \cite{iizuka2017globally} 12.13dB, that of \cite{nazeri2019edgeconnect} 15.66dB. Only when the tampered area is mostly at the background, the L-PSNR of the methods are close, namely, 26.97dB for Imuge, 24.77dB for \cite{iizuka2017globally} and 26.25dB for \cite{nazeri2019edgeconnect}. Therefore, Imuge content recovery is crucial for applications that require image fidelity.

\subsection{Ablation Tests}
We begin with implementing Imuge without the verifying network $\mathcal{V}$ and the discriminators. In this scenario, the attacked image is directly fed into the decoder, and Imuge has to recover the original content with the existence of tampered content at the same location. Then we introduce the verifying network $\mathcal{V}$. Afterwards, we introduce the discriminators, but during training we do not apply any training strategies. After that, we introduce task decoupling according to Section \ref{subsection:lossfunc}, where Imuge still tries to directly conduct content recovery in one go. Finally, the progressive recovery is further applied, but the local feature sharing is not used. The immunized images are attacked using the JPEG compression.

Table~\ref{Table3} provides the average performances of the above tests. We can observe that the experimental results gradually improve in PSNR within actual tampered areas. Specifically, in the first test, Imuge fails to conduct content recovery without identifying tampered regions. Although with the help of the verifying network, the tampers are removed, the scheme still produces messy contents within tampered areas. Then, the performances significantly improve with the introduction of the two discriminators. Finally, with the proposed mechanism, the best performance for Imuge so far can be achieved. The tests highlight the importance of the training mechanisms and the pipeline design.

\section{Conclusion}
\label{section:conclusion}
In this paper, we present a novel generative scheme called Imuge, which is an image tamper resilient generative scheme for image self-recovery. Imuge transforms the normal images into the immunized images, where the tamper attacks can be accurately localized and the original content within the tampered areas can be recovered. We train an encoder to produce a visually indistinguishable immunized version of a given image, a verifying network network to localize the malicious modifications, and a decoder to conduct the content recovery. Several training mechanisms are proposed to boost the training efficiency. The scheme is tested on natural images from different sources and the immunized images are mixed with a variety of complicated tampers and attacks. The experiments prove the effectiveness of Imuge in both tamper localization and content recovery.


\begin{acks}
This work is supported by National Natural Science Foundation of China under Grant U20B2051, U1936214.
\end{acks}

\bibliographystyle{ACM-Reference-Format}
\bibliography{reference}


\end{document}